\documentclass[letterpaper, 10 pt, conference]{ieeeconf}  

\usepackage{subfigure}
\IEEEoverridecommandlockouts                              

\usepackage{balance} 
\usepackage{amssymb,mathtools} 
\usepackage{gensymb} 
\usepackage{tabularx}
\usepackage{multirow} 
\usepackage{graphicx} 
\usepackage{booktabs,tabulary,lipsum}

\usepackage{caption}
\usepackage{tcolorbox}
\usepackage{tikz}
\usepackage{amsmath}

\usepackage[ruled,vlined, norelsize]{algorithm2e}
\let\labelindent\relax
\usepackage{enumitem} 

\title{\LARGE \bf
From Spoken Thoughts to Automated Driving Commentary: Predicting and Explaining Intelligent Vehicles' Actions
}

\author{Daniel Omeiza$^{1}$, Sule Anjomshoae$^{2}$, Helena Webb$^{3}$, Marina Jirotka$^{1}$, Lars Kunze$^{4}$
\thanks{$^{1}$Daniel Omeiza and Marina Jirotka are with the Dept. of Computer Science, University of Oxford. Email: {\tt\footnotesize daniel.omeiza@cs.ox.ac.uk}}%
\thanks{$^{2}$Sule Anjomshoae is with the Dept. of Computing Science, Umeå University, Sweden. Email: {\tt\footnotesize sule.anjomshoae@umu.se}}%
\thanks{$^{3}$Helena Webb is with the Dept. of Computer Science, University of Nottingham. Email: {\tt\footnotesize helena.webb@nottingham.ac.uk}}%
\thanks{$^{4}$Lars Kunze is with the Oxford Robotics Institute, Dept. of Engineering, Science,
        University of Oxford. Email: {\tt\footnotesize lars@robots.ox.ac.uk}}%
}
\SetKwInput{KwInput}{Input}
\SetKwInput{KwOutput}{Output}

\begin{document}

\maketitle
\thispagestyle{empty}
\pagestyle{empty}

\newcommand\copyrighttext{
  \footnotesize ©2022 IEEE. Personal use of this material is permitted. Permission from IEEE must be obtained for all other uses, in any current or future media, including reprinting/republishing this material for advertising or promotional purposes, creating new collective works, for resale or redistribution to servers or lists, or
reuse of any copyrighted component of this work in other works.}

\newcommand\copyrightnotice{
\begin{tikzpicture}[remember picture,overlay]
\node[anchor=south,yshift=10pt] at (current page.south) {\fbox{\parbox{\dimexpr\textwidth-\fboxsep-\fboxrule\relax}{\copyrighttext}}};
\end{tikzpicture}%
}

\begin{abstract}
In commentary driving, drivers verbalise their observations, assessments and intentions. By speaking out their thoughts, both learning and expert drivers are able to create a better understanding and  awareness of their surroundings. In the intelligent vehicle context, automated driving commentary can provide intelligible explanations about driving actions, thereby assisting a driver or an end-user during driving operations in challenging and safety-critical scenarios. In this paper, we conducted a field study in which we deployed a research vehicle in an urban environment to obtain data. While collecting sensor data of the vehicle's surroundings, we obtained driving commentary from a driving instructor using the think-aloud protocol. We analysed the driving commentary and uncovered an explanation style; the driver first announces his observations, announces his plans, and then makes general remarks. He also makes counterfactual comments. We successfully demonstrated how factual and counterfactual natural language explanations that follow this style could be automatically generated using a transparent tree-based approach. Generated explanations for longitudinal actions (e.g., stop and move) were deemed more intelligible and plausible by human judges compared to lateral actions, such as lane changes. We discussed how our approach can be built on in the future to realise more robust and effective explainability for driver assistance as well as partial and conditional automation of driving functions.

\end{abstract}
\section{Introduction}
Intelligent vehicles providing driver assistance and/or automated driving functionalities are required to monitor, predict, and assess driving behaviour to prevent accidents.
During their operation, interventions of the vehicle might include warnings as well as automated braking and/or steering.   
In challenging and critical driving scenarios, intelligent vehicles are likely to make decisions that are confusing to end-users~\cite{cunneen2019autonomous, chakraborti2020emerging}, e.g., unexpectedly initiating a lane change. As a way to assist end-users, and to establish trust, explanation provisions have been put forward~\cite{wiegand2020d, ha2020effects, omeiza2021not}.  
\begin{figure}
\centering
\includegraphics[width=\columnwidth]{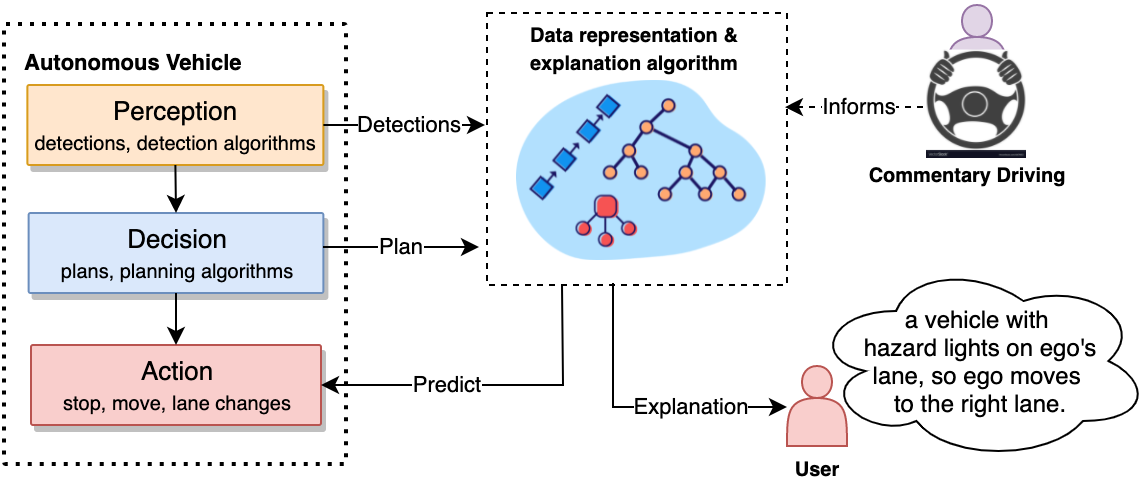}
\caption{\small{From commentary driving, requirements for explanations were gathered to inform the design of factual and counterfactual explanation algorithms. The algorithms receive input data from the different autonomous driving operations, provide a structured representation, and generate intelligible explanations to an end-user.}}
\label{fig:overview}
\end{figure}
While explanations are considered helpful, we argue that they would not be effective in achieving the aforementioned goals if they are not provided in intelligible forms as obligated by the General Data Protection Right (GDPR) Article 12\footnote{https://gdpr-info.eu/art-12-gdpr/}.

Intelligible explanations provision in assisted and automated driving is crucial as it is also a useful approach for upholding accountability. 
Intelligent vehicles should have explanation  mechanisms  where  the causes  and  effects  of actions  can  be  communicated  to  the relevant stakeholders in intelligible ways. 
Actions  resulting in accidents, undesired  outcomes, discriminatory,  and inequitable  outcomes need  to  be  accounted for.
This is important as accountability is part of the trust and confidence-building process.
A key question is how such explainability can be obtained in autonomous driving amidst the prevalence of complex models (`blackbox') for downstream driving tasks.

In this paper, we focus on unpacking the requirements for explanations in automated driving by analysing driving commentary~\cite{lunn1999commentary} and real-world observations. We describe a simple approach to implement some key requirements for explanation provision (see Figure~\ref{fig:overview}), and consequently suggest future considerations for more robust and effective implementations of explainability in intelligent vehicles.

Our main contributions are:
\begin{enumerate}

    \item explanation requirements elicitation based on spoken thoughts captured during real-world driving in an urban environment (Section~\ref{sec:fieldstudy});
    \item  a tree-based method with contextual importance estimation for generating driving commentary following the spelt out requirements (Section~\ref{sec:explanation});
    \item a quantitative and qualitative evaluation of the tree-based method and the generated explanations (Section~\ref{sec:experiments});
    \item discussion on future considerations for more robust implementation of explainability in IVs (Section~\ref{sec:discussion}).
\end{enumerate}  

\section{Related Work}
\label{sec:relwork}
Efforts have gone into the generation and communication of explanations for complex systems. Some of the approaches proposed are algorithm-based~\cite{chakraborti2020emerging}, that is, they investigate the underlying algorithm in detail to support debugging tasks~\cite{magnaguagno2017web}. These types of explanations are particularly tailored for specific algorithms (algorithm dependent) and the explainer is usually intrinsic. This means that the algorithm to be explained has built-in capabilities to generate explanations. Other explanation approaches are algorithm-agnostic in that they are able to assess the properties of a solution independent of the algorithms used to develop the solution~\cite{ribeiro2016should,lundberg2017unified}. A solution in this context is the behaviour of the agent. These explainers aim to help the inferential process of the end-user and also enhance the user's knowledge of the agent. Instead of explaining the deep learning algorithm behind the agent, our approach aims to explain the actions of the agent in natural language by observing and learning its behaviour (in a transparent way) over time.

In the bid to make explanations effective for different stakeholders, theories from the social sciences have been explored in line with explainable AI (XAI)~\cite{miller}.
Mittelstadt et al.~\cite{mittelstadt2019explaining} argued that the risk of conflicts in communicating explanations when the \textit{explainer} (explanation provider) and the \textit{explainee} (explanation recipient) have different motives may be mitigated through contrastive, selective, and social explanations. Social in the sense that the explanation process involves different parties and the explainer is able to model the expectations of the explainee. The explanation is selective if it is able to select explanations among several competing hypotheses. It is considered contrastive if it is able to differentiate the properties of two competing hypotheses. Boris Kment \cite{kment2006counterfactuals} also argued that counterfactuals are helpful in explaining. A counterfactual explanation describes reasons with respect to changes in input that lead to the change of a fact to a foil (competing hypothesis)~\cite{miller}. 
We investigated our research problems with these theories in mind.

Different methodologies have been adopted in the XAI literature. Wang et al.~\cite{wang2019designing} categorised these research methodologies into three groups:
First, Wang et al.~\cite{wang2019designing} highlighted the existence of \textit{unvalidated guidelines} for the design and evaluations of explanations. The authors in~\cite{wang2019designing} claimed that these kinds of guidelines are based on authors' experiences with no further substantial justification.
Second, researchers suggested (in~\cite{zhu2018explainable}) that understanding users' requirements is helpful in XAI research. It is on this premise that previous research on explanation design has been thought to be \textit{empirically derived}. This type of XAI research elicits explanation requirements from user surveys to determine the right explanation for a use-case with explanation interfaces.
Third, some explanation design methods are derived from \textit{psychological constructs from formal theories} in the academic literature. Some of these methods (e.g., in \cite{hoffman2017explaining}) draw on theories from cognitive psychology to inform explanation design for explanation frameworks. Our work is heavily grounded on empirical studies that applied the think-aloud methodology~\cite{van1994think}.

Closely related to explanations in AVs, recent works have applied deep learning approaches to explain actions in autonomous driving~~\cite{kim2018textual, xu2020explainable}. For example, Kim et al.~\cite{kim2018textual} proposed an approach for textual explanation generation for autonomous vehicles' actions through an attention-based video-to-text deep model trained on their BDD-X dataset. The explanations were provided in form of attention maps and texts.
These works are limited in terms of the transparency of the explainer, and the intelligibility of the explanations provided. Some authors advocate for interpretable approaches to explainability in AVs. For instance, Omeiza et al.~\cite{omeiza2021towards} proposed a social-technical approach to explainability and proposed an interpretable representation for explanations based on a combination of actions,  observations, and road rules. Nahata~et al.~\cite{nahata2021assessing} also applied an interpretable method to explain risk prediction models for autonomous vehicles.
Other recent research~\cite{schneider2021explain, omeiza2021not, koo2015did, ha2020effects} have explored the relationship between explanations and trust in AVs.

We build on the existing work to provide a socio-technical and more transparent approach to provide intelligible explanations that can offer counterfactual inferences while respecting set constraints.

\section{Field Study: Commentary Driving}
\label{sec:fieldstudy}
We conducted a field study to study how an expert driving instructor explains driving events following a think-aloud protocol, then demonstrate the possibility of designing an explainer that fulfils key observed requirements.

\subsection{Data Collection}
We deployed an ego vehicle to an urban environment with an expert driving instructor who drove the vehicle via different routes in London to collect data for our study. The driving instructor had trained to become an advanced driving instructor in the United Kingdom and had previous experiences using the spoken-thought technique while driving. We asked the driver to provide commentaries while he drove. The ego vehicle was fitted with different sensors among which were: microphones, cameras, and lidar. The in-cabin microphone recorded the commentaries, while the external cameras recorded the scenes in the environment. We annotated all the videos from the field trial with bounding boxes, high-level semantic, structured explanations (i.e., actions and causes/reasons), and the transcribed comments. In this study, we only selected three 1-hour long driving videos for driving commentary analysis. The video has different scenes: 87 stops, 31 right lane changes (RLC), 29 left lane changes (LLC), and many straight move actions in total.

\subsection{Analysis and Observations}
\label{sec:convo_analysis}
We analysed the video data with the aim to
(1) uncover explanation styles and requirements in real-world driving;
(2) inform future designs of explanation systems in AVs.

\subsubsection{Explanation Styles and Requirements}
We collected major stimulus-driven action~\cite{ramanishka2018toward} (the stop, move, right lane change, and left lane change) instances from each of the videos. Lane changes include right manoeuvres to a right adjacent lane and left manoeuvres to a left adjacent lane. We ensured that all comments of the driver related to these actions were part of the collected video instances; see Figure~\ref{fig:exp_imgs} for sample frames from the videos. We watched the video instances repeatedly for the four actions, we discovered that the driver in most cases made explanations in the following order: (i) announce observations, (ii) announce plans (with or without reference to road rules), and (iii) make general remarks.

\begin{enumerate}[label=(\roman*)]
    \item Announce observations: The driver first announces an observation within his limit points. The limit point is the farthest point along a road to which a driver has a clear and uninterrupted view of the road surface. When there are many observations, he \textit{selects and focuses} on the most interesting and relevant ones. 
    
    When describing an observation, the driver mentions the type of agent, the current action of the agent, and the road position of the agent. 
    \item Announce plans: Following the announcement of observations, the driver announces his planned action, and in some cases, how the plan respects the relevant road rule. In some cases, he states his plan during or after execution.
    \item General remarks: When the driver completes the execution of his initial plan and the vehicle is steadily moving, he starts making a general comment about special observations in his view, e.g., road topology, special architectural designs, and pedestrians' movement on the sidewalks.
\end{enumerate}

Excerpt from the description of an observation:
\begin{quote}
    \textit{Driver (Video 3: 09:45 -- 09:53):} `We've got a vehicle on the left part indicating so lights are on so he's live, we'd leave space so if he does move out [paused] now we'll pull back into the lane'\\
    \textit{Driver (Video 2: 24:14 -- 24:21):} `Got a bus pulling up on the left hand side and we've got lots of bright lights so there's gonna be traffic coming up there.'
\end{quote}

The amount of space an agent in front of the driver occupied affected the driver's decision. The scene became an interesting one to comment on.
\begin{quote}
    \textit{Driver (Video 1: 24:52 -- 25:13)}: `I could see but I'm quite a big vehicle. And by the time I've sort of gone in and gone out he's probably cleared anyway...Again, if I were on a little sports car, I might do it. But this is just a little bit bigger, I've got to be a bit more cautious with it.'
\end{quote}

In many instances, the driver made counterfactual inferences:
\begin{quote}
\textit{Driver (Video 1: 06:55 -- 06:59): } `If I tried to go around the bus, I'd have been strained to the back of a vehicle that was already there.'
\end{quote}

Generally, the most common causes for the driver's stop actions were traffic lights, pedestrians crossing, and static traffic queues.
The common reasons for lane changes were to walk around a parked vehicle, overtake a vehicle, move to a faster lane, or return to its default lane.

For the rest of this paper, we aim to demonstrate how intelligible natural language explanations can be generated for an AV's action, by describing observations (stating agent type, action, position) and plans as observed in the study. We will show how explanations could be selective in the presence of competing causes. We will also demonstrate the generation of counterfactual explanations. To do this, we first formalise relevant terms for an easy description of our method.






\section{Preliminaries}
In the context of this paper, we consider an autonomous vehicle $a_v \in \mathcal{A}$ as a special type of agent (i.e., a driverless car) in a shared environment. $\mathcal{A}$ is a set of agents of different classes. Along with $a_v$, other agents (say $ a_{i} \in \mathcal{A}$) also exist in the shared environment.

Each $a$ (including $a_v$ and $a_{i}$) has information $Y \subseteq \mathcal{Y}$ indicating its class ($C_a$), action ($\mathcal{X}_a$) and position ($\mathcal{P}_a$) at a given time $t \in \mathcal{T}$.  So, ${Y_t}: {Y_t} = \{{C_{a}}, {\mathcal{X}_{a}}, {\mathcal{P}_{a}}\}$. Time is a real number $\mathcal{T} = \mathbb{R}^{+}$. 
$a_v$ needs to observe other agents  to plan its trajectory while respecting certain constraints (e.g. road rules and restrictions). AV's planned trajectory obtained at time $t \in \mathcal{T}$ is denoted as ${\xi_{a_v}(t)}$.
$\mathcal{M}$ denotes a tree-based model that receives feature vector $V$ as input and predicts the AV's current action $\mathcal{X}_{a_v}$. $V$ may contain information $Y_t$ of other relevant agents in the shared environment, and the AV's planned trajectory ${\xi_{a_v}(t)}$.

In this paper, we represent actions ($\mathcal{X}_a$) and trajectories (${\xi(t)}$) as high-level SatNaV commands e.g., move, stop, lane change, among others. $\mathcal{H}$ is used to denote the set of constraints for generating counterfactual explanations for the AV's actions.

\section{Problem Formulation}
Generally, our goal is to predict $\mathcal{X}_{a_v}$ given $V$ and generate intelligible explanations $E$ for the prediction. For $E$ to be intelligible, it should make references to the influential features in $V$ using clear natural language semantics.

Following the requirements in Sections~\ref{sec:relwork} and \ref{sec:convo_analysis} we want to be able to: (i) generate clear factual (or why) explanations
(ii) select the relevant factual explanation for action amidst different competing factual explanations.
(iii) generate clear counterfactual (or what-if) explanations while respecting some constraints $\mathcal{H}$. In our case, constraints are restrictions in the form of input features whose attributes should not be modified when generating a counterfactual explanation.

\section{Explanation Generation Approach}
\label{sec:explanation}
We propose the use of tree-based models  which are easily explainable to learn the behaviour of an AV from our annotated video dataset. In particular, tree-based models avail the opportunity to inspect decision boundaries and split criteria at every decision node. This makes the generation of intelligible natural language explanations possible. Similar approaches have been used in~\cite{nahata2021assessing} for collision risk assessments, and in \cite{stepin2021factual} for explaining fuzzy systems. These works are limited as they do not implement requirements for cause selection from a set of competing causes and do not also implement constraints for generating counterfactuals. Simply put, in the context of the tree-based approach, a cause is a split condition for a feature at a node. We apply the notion of contextual importance (CI). CI expresses the importance of the different feature attributes for a prediction. Apart from being important, we want to know the extent to which the attributes of the different input features are favourable (or not) for a prediction, this is referred to as contextual utility~\cite{anjomshoae2021context}.

In Algorithm~\ref{alg:tb}, we pass as input a tree model $\mathcal{M}$ which has been trained on a dataset containing records of encoded information ${Y}_t$ for different agents in a shared environment at time $t$. We also pass the AV action predicted by $\mathcal{M}$. $\mathcal{M}$ can be a classification or regression decision tree, and can also be an ensemble of such trees.

\begin{algorithm}
\LinesNumbered
\DontPrintSemicolon
\KwInput{tree model $\mathcal{M}$, input vector $V$, predicted AV's action $\mathcal{X}_{a_v}$}
\KwOutput{intelligible textual factual explanation}
    $causes \longleftarrow \emptyset$\\
    $ci \longleftarrow obtainCI(\mathcal{M}, V)$\\
    \If {$ isEnsemble(\mathcal{M})$}{
        $\mathcal{X}_{path} \longleftarrow \emptyset$\\
        \If{isClassification($\mathcal{M}$)}{
            $\mathcal{M} \longleftarrow obtainModeTrees(\mathcal{M}, V)$\\
            \For{$m \in \mathcal{M}$}{
            $\mathcal{X}_{path} \longleftarrow \mathcal{X}_{path} \cup obtainDecPath(m, V)$
            }
            $\mathcal{M} \longleftarrow obtainTreebyFactoringPaths(\mathcal{M}, \mathcal{X}_{path})$\\
        }
        \uElseIf {$ isRegression(\mathcal{M}) $}{
            $\mathcal{M} \longleftarrow obtainMedianTree(\mathcal{M}, V)$
        }
    }
    $\mathcal{X}_{path} \longleftarrow obtainDecPath(\mathcal{M}, V)$\\
    $causes \longleftarrow mergeInequilities(\mathcal{X}_{path})$ \\
    $selected \leftarrow obtainRelevantCauses(ci, causes)$\\
    $entropy \leftarrow Entropy(\mathcal{M}, V)$\\
    \Return decode($\mathcal{X}_{a_v}$, $selected$, $\mathcal{X}_{path}$, $entropy$)
\caption{Tree-based Factual Explanation}
\label{alg:tb}
\end{algorithm}




In the tree-based algorithm, we first obtain the CI values for the feature attributes in $V$ (line 2).
We check whether $\mathcal{M}$ is a random forest model. We further check whether $\mathcal{M}$ is a classification model or regression model. If $\mathcal{M}$ is a random forest classifier, we obtain all the trees in the forest that predicted the action $\mathcal{X}_{a_v}$ (line 6). Subsequently, the paths from the root to the $\mathcal{X}_{a_v}$ node of the resulting trees are obtained and the tree that contains the most re-occurring features across these trees is selected (lines 7 - 9).
If the model is a random forest regressor, we obtain the tree that predicted the closest value to the median of all predicted $\mathcal{X}_{a_v}$ (lines 10 - 11). 
The decision path for the resulting classification or regression tree is obtained and the feature conditions along this path are merged to have a single decision boundary for each feature (lines 12 - 13) as there can be many split conditions for each feature. We refer to these merged conditions as causes. For example, if we have two conditions in a path: $\{\{`Feature1' < 50\}, \{`Feature1' < 10\}\}$, the merged path will be ${\{`Feature1' < 10\}}$.

We then look for the causes whose features have high positive CIs (line 14). We do this by obtaining the cause with the maximum CI, and then adding more causes if the percentage difference between their CIs and the maximum CI is less than a threshold (we used a threshold of 50\%).

We estimate the level of uncertainty of the explainer by computing the information entropy of the training sample distribution in the leaf node of the decision path (line 15). This is obtained by:
\begin{equation}
S := \sum_{i = 1}^N {-p_i log_2 p_i}
\end{equation}
While estimating CI (in line 2), we explored tree specific methods (local increments \cite{palczewska2013interpreting} and Tree SHAP~\cite{lundberg2017unified}) that could provide this information. We chose Tree SHAP algorithm as it provided the best accuracy. Tree SHAP estimates the SHAP values of each feature \(i\) from \(1,...,N\). The fundamental procedure is given as:

\begin{itemize}
    \item generate all subsets \(S\) of the set $F = \{1, ... , N \setminus\{i\}\}$
    \item for each \(S \subseteq F\setminus\{i\}\) estimate the contribution of feature \(i\) as \(CT\{i|S\}~=~\mathcal{M}(S\cup\{i\})-\mathcal{M}(S)\)
    \item compute the SHAP value according to:
\end{itemize}

\begin{equation}
\phi_i := \frac{1}{N}\sum_{S \subseteq F\setminus\{i\}}{\binom{N-1}{|S|}}^{-1}CT(i|S)
\end{equation}
High positive SHAP values indicate high importance and utility, while very low negative SHAP values indicate high importance but low utility. The combination of CIs and the conditions along the decision path allows for the provision of more specific and intelligible explanations. The decode function (line 16) provides the human-understandable textual explanation based on the supplied values. 

If a counterfactual explanation is desired, a corresponding counterfactual explanation is generated using Algorithm~\ref{alg:tbcf}.
\begin{algorithm}
\DontPrintSemicolon
\LinesNumbered
\KwInput{tree model $\mathcal{M}$, input vector $V$, predicted AV's action $\mathcal{X}_{a_v}$, expected counterfactual output $\mathcal{X}_{a_v}^{\prime}$, constraints $\mathcal{H}$}
\KwOutput{intelligible textual counterfactual explanation}
$\mathcal{X}_{cfpath} \longleftarrow \emptyset$\\
$n_a \longleftarrow \emptyset$\\
    \If{${\mathcal{X}_{a_v}^{\prime}} \equiv \emptyset$}{
        ${\mathcal{X}_{a_v}^{\prime}}  \longleftarrow findClosestCFSibling(\mathcal{M},V, {\mathcal{X}_{a_v}}, {\mathcal{H}})$\\
    }
    $n_a, \mathcal{X}_{cfpath} \longleftarrow lowestCommonAncestor(\mathcal{M},V,  {\mathcal{X}_{a_v}}, {\mathcal{X}_{a_v}^{\prime}}, \mathcal{H})$\\

    $\mathcal{X}_{cfpath}[n_a] \longleftarrow \neg \mathcal{X}_{cfpath}[n_a]$ \\
    $conditions \longleftarrow mergeInequilities({\mathcal{X}_{cfpath}})$\\
    $entropy \leftarrow Entropy(\mathcal{M}, V, {\mathcal{X}_{a_v}})$\\
    \Return decode(${\mathcal{X}_{a_v}^{\prime}}$, $conditions$, $\mathcal{X}_{cfpath}$, $entropy$)
\caption{Tree-based Counterfactual Explanation}
\label{alg:tbcf}
\end{algorithm}
To construct a counterfactual explanation, Algorithm~\ref{alg:tbcf} first checks whether a desired counterfactual output (${\mathcal{X}_{a_v}^{\prime}}$) was provided (line 3). When not provided, it finds the closest sibling node (${\mathcal{X}_{a_v}^{\prime}}$) to the leaf node (${{\mathcal{X}_{a_v}}}$) subject to the constraint that ${{\mathcal{X}_{a_v}}} \neq {\mathcal{X}_{a_v}^{\prime}}$ (lines 3 - 4).
Another type of constraint used is such that restricts the modification of a feature attribute while searching for counterfactual candidates.

\begin{figure}[htb!]
\centering
\includegraphics[width=\columnwidth]{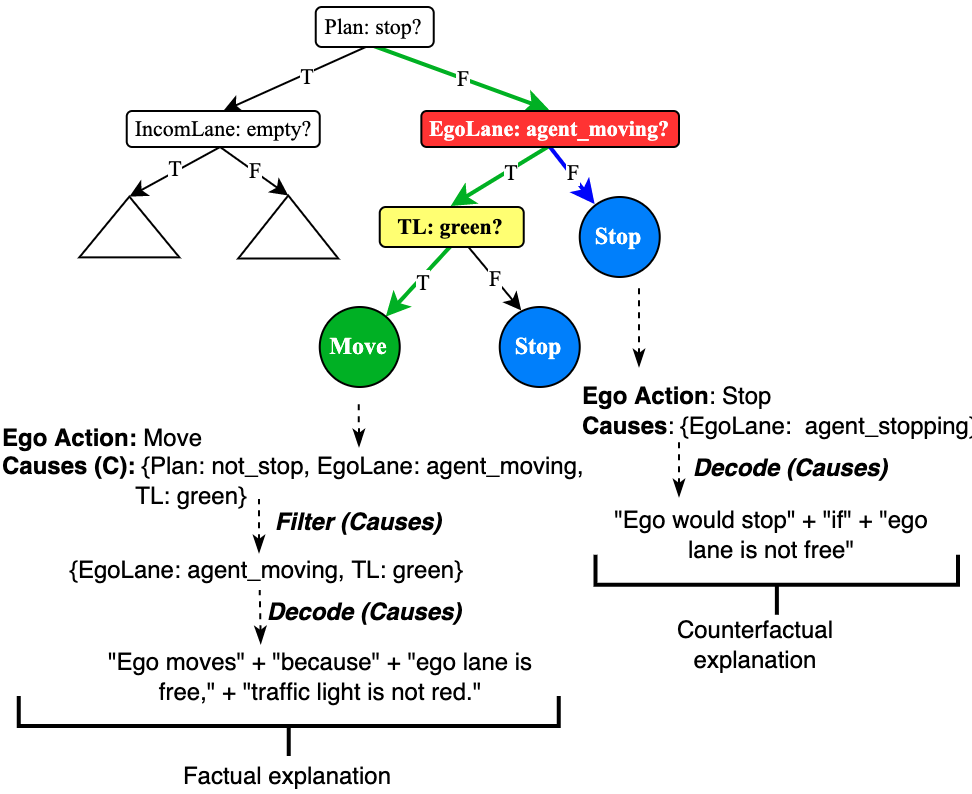}
\caption{\small Explanation generation process. The green edges are the model's decision path for a move action prediction. The conditions/causes on the decision path are filtered based on CI values. The decode function provides the natural language texts based on a predefined mapping of node conditions to English phrases. The blue nodes are the counterfactual candidates. The yellow node has a feature condition that is non-modifiable based on a set constraint. The closest sibling node to the predicted move action is the blue node below the yellow node, but because there is a constraint on the feature in the yellow node, we move a level up the tree to find the next closest sibling which is the blue node below the red node. The selected counterfactual candidate is the blue path from the red node to this new sibling node (rightmost blue node). The condition in the red node is negated and the resulting list of conditions/causes is decoded to form a natural language counterfactual explanation.}
\label{fig:tree_overview}
\end{figure}

When $ {\mathcal{X}_{a_v}^{\prime}} $ is provided, the algorithm only finds the lowest common ancestor of ${{\mathcal{X}_{a_v}}}$ and  ${\mathcal{X}_{a_v}^{\prime}}$ obtaining the counterfactual conditions in the process (line 5). 
The condition at node $n_a$ is negated and $\mathcal{X}_{cfpath}$ is updated to only contain counterfactual conditions (line 6). See 
Figure~\ref{fig:tree_overview}.
\section{Experiments}
\label{sec:experiments}
As shown in Figure~\ref{fig:overview}, we used insights from the field study to design an explanation algorithms that predict an ego vehicle's action based on detections/observations and high-level plans and then provide an intelligible explanation for the action predicted. We first explain how we annotated the dataset on which $\mathcal{M}$ was trained, and then explain the properties of $\mathcal{M}$. 

\subsection{Dataset Annotation}
In our annotation processes, we made the best effort in annotating different driving scenarios with a mix of objective criteria and subjective judgment. Our annotations scheme is similar to that used in~\cite{singh2021road}.
An event in a scenario, among other interesting information, comprises agents of various classes $C_{a_i}$ (e.g., vehicle, motorbike, pedestrian, traffic light), their actions $\mathcal{X}_{a_i}$ (e.g., moving, crossing), locations relative to the AV $\mathcal{P}_{a_i}$ (e.g., Ego lane, incoming lane, outgoing lane), plan/trajectory ${\xi_{a_i}}(t)$ (e.g., move, stop), influence on the AV (e.g., primary influence), and the driver's comment on the scene at the time. We generated the ground truth explanations for the ego actions based on the influence tags that were added to the agents that influenced the ego's actions. We converted this semantic information into tabular forms where we have the following columns as features to our tree model: \textit{EgoLane, IncomLane, OutgoLane, TL, EgoPlan, and EgoAction}. \textit{EgoLane} is the current lane that the ego vehicle travels on. The \textit{IncomLane} is an adjacent lane on which traffic flows in the opposite direction to the ego vehicle's direction. The \textit{OutgoLane} is an adjacent lane in which traffic flows in the same direction as the ego vehicle's direction. Most of the relevant agents on the road fall within one of these lanes. \textit{TL} is traffic light, and \textit{EgoPlan} is the ego vehicle's high-level trajectory which is simply the next action (that is n-frames ahead). Agents with their corresponding actions were numerically encoded to serve as attributes for these features. Where there is more than one agent on a lane, following the human driver approach from our field study, we selected the most dominant one. That is the one that influences the ego vehicle's decision the most. This is dependent on the size and proximity of the agent; this also includes pedestrians. Each record in our dataset represents a frame. We randomly sampled our training and test sets from this dataset. However, there were not as many lane change examples as stops and moves due to the rarity of such events. The stop and move actions dominate the dataset. Moreover, the road topology in London, the ego's goal, the presence of several traffic jams and traffic lights are contributing factors to the higher frequency of the stop and move actions.

\subsection{The Explainer Model}
 We trained a decision tree model and a random forest model on the curated dataset. We choose the random forest model as ($\mathcal{M}$) in the experiment as it yielded higher accuracy for our task. There were a total of $2,755$ sampled records from the original dataset, with 800 stop instances, 900 move instances, 483 right lane change instances, and 572 left lane change instances of the AV. The train/test split ratio was 80:20. 
 We fitted a random forest tree model using 10 folds cross-validation. Given a driving video, the tree-based model can be applied at each instance $t \in \mathcal{T}$ to predict the ego's action which is subsequently explained using the described explanation algorithms.

\begin{figure}[htb!]

\centering
\includegraphics[scale=0.6]{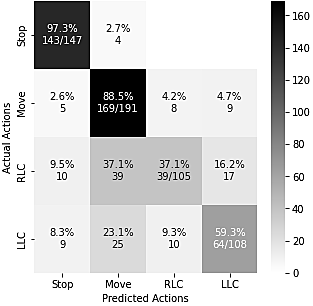}
\caption{\footnotesize The model yielded a test accuracy of 0.75 with higher performance for stop and move actions.}
\label{fig:conf_mat}
\end{figure}

The model yielded a test accuracy of 0.75 with higher performance for the stop and the move actions. This model $\mathcal{M}$ was used in Algorithms~\ref{alg:tb} and \ref{alg:tbcf} to generate explanations.
See more detail about the model's performance in Figure~\ref{fig:conf_mat}.

\begin{figure*}
\centering
\includegraphics[width=\linewidth]{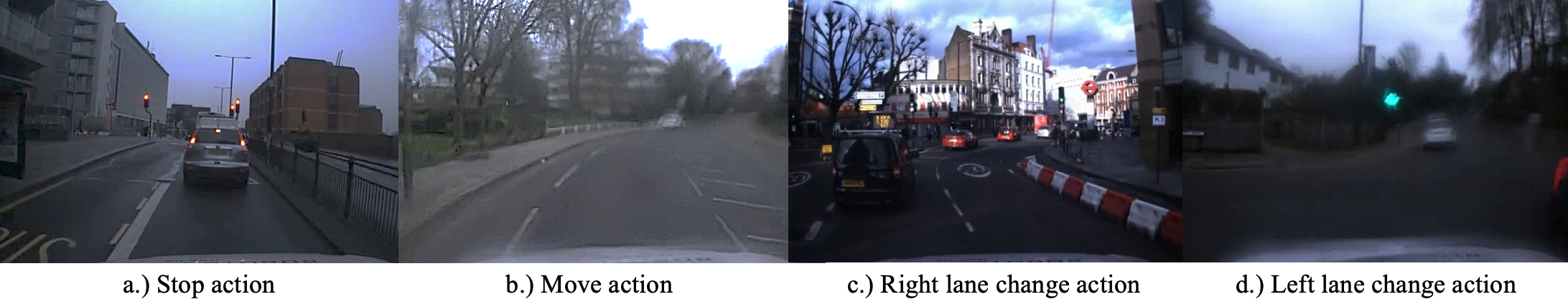}
\small
\begin{tcolorbox}[colframe=gray,colback=white,boxrule=2pt,arc=0.6em,boxsep=-1mm]
\footnotesize{
\textbf{a.) Stop action explanation:} \\
\textit{Factual Explanation:} ``Traffic light is not green on ego's lane, so ego stops" $S=0.45$\\
\textit{Counterfactual Explanation:}
''If ego must move straight, the following should be happening: road is free on ego's lane; the traffic light is green on ego's lane'' $S=0.46$

\textbf{b.) Move action explanation:} \\
\textit{Factual Explanation:} ``Vehicle is moving on ego's lane, so ego moves" $S=0$\\
\textit{Counterfactual Explanation:}
''If ego must stop, the following should be happening: a vehicle stopped on ego's lane'' $S=0$

\textbf{c.) Right lane change action explanation:} \\
\textit{Factual Explanation:} ``Traffic light is green on ego's lane, so ego moves straight" $S=0.83$\\
\textit{Counterfactual Explanation:}
''If ego must move straight, the following should be happening: a vehicle stopped on outgoing lane'' $S=0.83$

\textbf{d.) Left lane change action explanation:} \\
\textit{Factual Explanation:} ``Ego's next goal is to move to the right lane, so ego moves to the left lane" $S=0$\\
\textit{Counterfactual Explanation:}
''If ego must move to the right lane, the following should be happening: a vehicle is braking on ego's lane'' $S=0.65$}

\end{tcolorbox}
\caption{\footnotesize
Sample factual and counterfactual explanations for the four actions along with entropy scores: (a) Ego stops due to traffic light; (b) Ego moves as a vehicle in front moves; (c) Right lane change action is misclassified as a move action, and thus its explanation would better be suited for a move action; (d) Explanation is based on a plan. Ego moves to the left lane and then plans to counteract this by moving to the right lane. Explanations with lower entropy seem to be slightly more plausible.}
\label{fig:exp_imgs}
\end{figure*}

\section{Results}
We first provide examples of scenarios in which our explainer model was applied to generate explanations.
\subsubsection{Scenario A} In the stop scene depicted in Figure~\ref{fig:exp_imgs}a, the ego vehicle was stopping in front of a vehicle in a stop state due to the red traffic light. The explainer selected only the traffic light as the cause for the ego vehicle's stop action; therefore fulfilling the selective requirement. The factual explanation follows the sequence uncovered in Section~\ref{sec:convo_analysis}, i.e., \textit{announce observation $\rightarrow$ announce plan/action}. A counterfactual explanation is also generated. In this case, the desired counterfactual action is \textit{move} based on the plan of the ego vehicle. In this example, we placed a constraint on the EgoPlan feature so that the ego's plan is not modified when generating counterfactuals.

\subsubsection{Scenario B}
The scene in Figure~\ref{fig:exp_imgs}b depicts a move action. The ego vehicle keeps moving as long as the vehicle ahead moves. Its future plan is a stop, according to the counterfactual explanation this would happen if a vehicle stops in front of the ego on the ego's lane.

\subsubsection{Scenario C} 
In Figure~\ref{fig:exp_imgs}c, a right lane change action was misclassified as a move action. The entropy is also the highest in these examples, so we put less confidence in the explanations in this example.
\subsubsection{Scenario D} 
Figure~\ref{fig:exp_imgs}d depicts a left lane change action. The ego provided a factual explanation based on its next plan. While the counterfactual explanation might be plausible, the current ego lane runs into roadside buildings so ego had to change lane to the right as soon as possible. This limitation occurred as we did not consider off-road objects and static objects in our study.

\begin{figure}
\centering
\includegraphics[height=3.5cm, width=7.3cm]{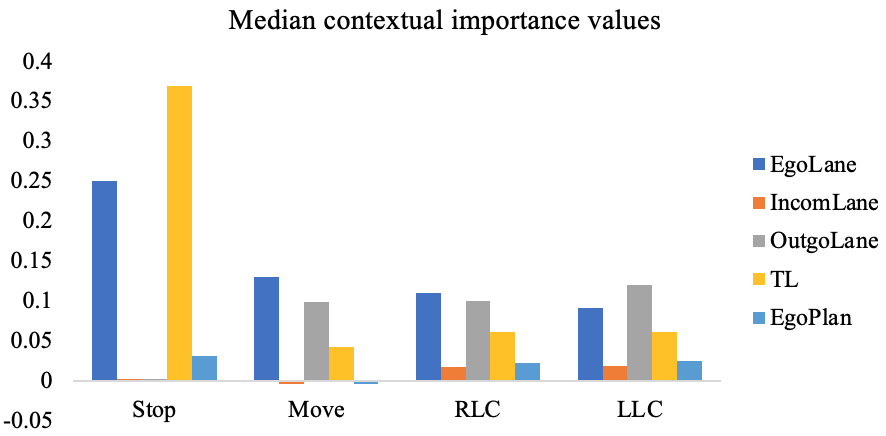}
\caption{\footnotesize Overall, traffic lights, observations on the vehicle lane, and observations on the outgoing lane had the highest contribution to the actions of the ego vehicle.}
\label{fig:cicu}
\end{figure}

Finally, we computed the median contextual importance scores (CI) scores for the correct prediction from all the test sets for the four actions; see Figure~\ref{fig:cicu}.
Overall, traffic lights, observations on the vehicle lane, and observations on the outgoing lane had the highest contribution to the actions of the ego.

\subsection{Quantitative Results}
\begin{figure}[htb!]
\centering
\includegraphics[height=4.7cm, width=7.4cm]{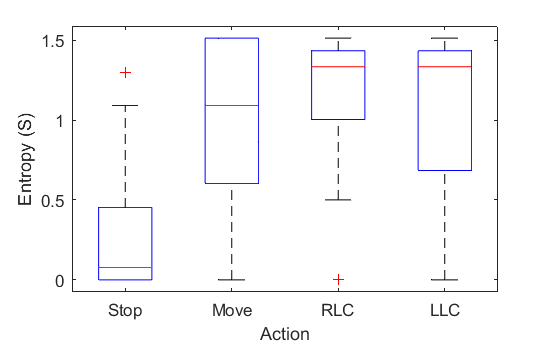}
\caption{\footnotesize Certainty of the explainer per class with overall minimum = 0, overall maximum = 1.46, overall median = 0.95.}
\label{fig:certanity}
\end{figure}

We measured the amount of similarity between the generated factual explanations with ground truth explanations using the BiLingual Evaluation Understudy (specifically cumulative weighted BLEU-4) and The Recall-Oriented Understudy for Gisting Evaluation (specifically the weighted LCS ROUGE-W). The mean similarity score is calculated based on the averaged median entropy (0.95) for each class. The lower entropy means that the model has higher certainty on the generated explanation. Figure~\ref{fig:certanity} shows the distribution of the entropy values per class. As the figure shows, the model had the highest certainty on the explanations generated for stop actions.

In terms of similarity scores, there was a slight increase considering the model’s certainty (See Table~\ref{tab:certanity}). Overall, stop and move action gave the highest similarity scores for both matrices.  

{\fontsize{8pt}{11pt}
\selectfont
\begin{table}[htb!]
\centering
\begin{tabular}{lccccl}\toprule
& \multicolumn{2}{c}{BLEU-4} & \multicolumn{2}{c}{ROUGE-W}
\\\cmidrule(lr){2-3}\cmidrule(lr){4-5}
           & $ S \leq .95 $  & $S > .95$ & $ S \leq .95 $  & $ S> .95$   \\\midrule
Stop    & .648 & .561 & .732 & .653  \\
Move & .537 & .693 & .763 & .783 \\
RLC & .594 & .348 & .697 & .462 \\
LLC & .498 & .568 & .627 & .672 \\\bottomrule
\end{tabular}
\caption{\footnotesize Comparing generated factual explanations with ground truth explanations. We chose a median entropy value of .95. Similarity scores seemed to slightly increase with lower entropy values. $Min(S) = 0, Max(S)=1.46, Median(S) = 0.95.$
}
\label{tab:certanity}
\end{table}
}
\subsection{Qualitative Results}
\begin{figure}[htb!]
\centering
\includegraphics[width=\columnwidth]{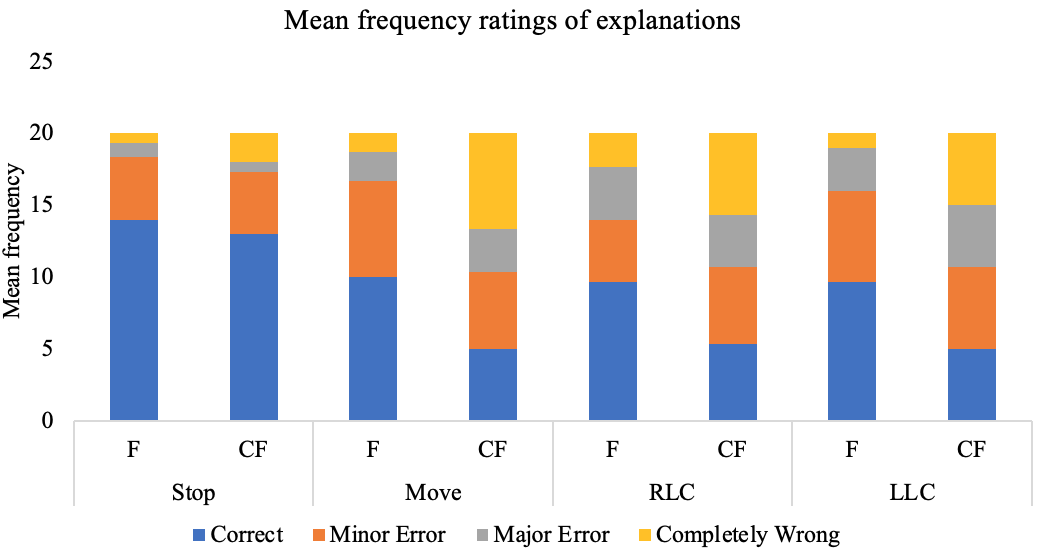}
\caption{\footnotesize Participants' ratings for the factual and counterfactual explanations for each class.}
\label{fig:ratings}
\end{figure}

We randomly selected twelve 4-second videos. We ensured that each ego action had 3 examples. We presented these videos with their corresponding generated explanations to 20 human judges (10 males and 10 females) all with driving licences and driving experiences in the UK. These judges were recruited through the Prolific platform and were between the age of 18 and 65. We asked the participants to rate both the factual and counterfactual explanation on the scale {0...3} (3: correct, 2: minor error, 1: major error, 0: completely wrong). Factual and counterfactual explanations for stop actions had the highest ratings for correct explanations (mean frequency of 14 and 13 respectively). The lowest rated factual and counterfactual explanations were those for right lane change (mean frequency of 2.333) and move actions (mean frequency of 6.667) respectively (See Figure~\ref{fig:ratings}).

\section{Discussion}
\label{sec:discussion}
The proposed approach is simple, has increased transparency, and supports the generation of specific causes for effects in driving rather than just feature importance scores that other explanation algorithms provide. This research, in general shows the potential for the realisation of automated explainers that can provide driving commentary to help the inference process of end-users in the driving context and also increase the driving knowledge of learner drivers. However, there are limitations in the current work; the issues pointed out in Scenarios C and D in our experiment are due to the limited size of the collated dataset, especially for the lane change classes. Model performance could be promising with more examples of the rarer classes. Moreover, the consideration of relevant static road elements might be useful for achieving a more robust explainer.

Apart from the insights discussed in Section~\ref{sec:convo_analysis}, there are other insights to learn from how humans explain events, for the benefit of future research. For example, early on, during the ride in the field study, the driver provided basic information about his driving preference. He stated his lane preference and his use of the side mirrors. We suggest that the driving preferences programmed into AVs be provided to the passengers at the start of a ride. It is important for an AV to be able to detect special vehicles e.g., foreign trucks, and anticipate their actions as the driver did throughout the driving exercise. Explanations should therefore be able to reference these subtle differences. As some of the driver's comments reflect that he made decisions based on other agents' distances, relative distance metrics between an AV and other road agents may be relevant for explanation purposes. Although challenging, we think explainers should model how humans explain~\cite{de2017people}, while also noting the difficulty experts might have in explaining their actions~\cite{nisbett1977telling, kahneman2009conditions}.
While high-fidelity human-centred studies with stakeholders in the loop may be expensive and challenging to set up in autonomous driving research, they are very useful for eliciting requirements and learning preferences to inform the development of more robust explainer systems.
In future work, we will strengthen the proposed explanation technique to be more robust and allow for the explanation of more fine-grained actions and observations.

\section{Conclusion}
\label{sec:conclusion}
In this paper, we explored the requirements needed for an effective explanation of driving actions in order to gain insights into the design of a more transparent explanation method for intelligent vehicles' actions. The long-term goals are to support in-vehicle participants, increase accountability, foster confidence and trust in autonomous vehicles. We provided a transparent approach that can explain the stop, move and lane change actions of an ego vehicle based on a driving dataset that we collected. Results from our experiment indicate the possibility of explaining the actions of AVs in the real world. With further work, this solution can be deployed in intelligent vehicles or AVs in the future.

\section*{ACKNOWLEDGMENTS}
This work was supported by the Assuring Autonomy International Programme (Demonstrator project: Sense-Assess-eXplain (SAX)), a partnership between Lloyd's Register Foundation and the University of York. It was also supported by the EPSRC projects RoboTIPS (grant reference: EP/S005099/1), RAILS (grant reference: EP/W011344/1), and RoAD (Responsible AV Data) grant reference: EP/V00784X/1.

\footnotesize{
\bibliographystyle{ieeetr}
\bibliography{bibliography}

\begin{thebibliography}{10}

\bibitem{cunneen2019autonomous}
M.~Cunneen, M.~Mullins, and F.~Murphy, ``Autonomous vehicles and embedded
  artificial intelligence: The challenges of framing machine driving
  decisions,'' {\em Applied Artificial Intelligence}, vol.~33, no.~8,
  pp.~706--731, 2019.

\bibitem{chakraborti2020emerging}
T.~Chakraborti, S.~Sreedharan, and S.~Kambhampati, ``The emerging landscape of
  explainable automated planning \& decision making.,'' in {\em IJCAI},
  pp.~4803--4811, 2020.

\bibitem{wiegand2020d}
G.~Wiegand, M.~Eiband, M.~Haubelt, and H.~Hussmann, ``“i’d like an
  explanation for that!” exploring reactions to unexpected autonomous
  driving,'' in {\em 22nd International Conference on Human-Computer
  Interaction with Mobile Devices and Services}, pp.~1--11, 2020.

\bibitem{ha2020effects}
T.~Ha, S.~Kim, D.~Seo, and S.~Lee, ``Effects of explanation types and perceived
  risk on trust in autonomous vehicles,'' {\em Transportation research part F:
  traffic psychology and behaviour}, vol.~73, pp.~271--280, 2020.

\bibitem{omeiza2021not}
D.~Omeiza, K.~Kollnig, H.~Webb, M.~Jirotka, and L.~Kunze, ``Why not explain?
  effects of explanations on human perceptions of autonomous driving,'' in {\em
  2021 IEEE International Conference on Advanced Robotics and its Social
  Impacts}, IEEE, 2021.

\bibitem{lunn1999commentary}
B.~Lunn, ``Commentary driving: A tool for excellence,'' {\em Driver/Education},
  vol.~9, no.~3, 1999.

\bibitem{magnaguagno2017web}
M.~C. Magnaguagno, R.~FRAGA~PEREIRA, M.~D. M{\'o}re, and F.~R. Meneguzzi, ``Web
  planner: A tool to develop classical planning domains and visualize heuristic
  state-space search,'' in {\em 2017 Workshop on User Interfaces and Scheduling
  and Planning (UISP@ ICAPS), 2017, Estados Unidos.}, 2017.

\bibitem{ribeiro2016should}
M.~T. Ribeiro, S.~Singh, and C.~Guestrin, ``" why should i trust you?"
  explaining the predictions of any classifier,'' in {\em Proceedings of the
  22nd ACM SIGKDD international conference on knowledge discovery and data
  mining}, pp.~1135--1144, 2016.

\bibitem{lundberg2017unified}
S.~Lundberg and S.-I. Lee, ``A unified approach to interpreting model
  predictions,'' {\em arXiv preprint arXiv:1705.07874}, 2017.

\bibitem{miller}
T.~Miller, ``Explanation in artificial intelligence: Insights from the social
  sciences,'' {\em Artificial Intelligence}, vol.~267, pp.~1--38, 2019.

\bibitem{mittelstadt2019explaining}
B.~Mittelstadt, C.~Russell, and S.~Wachter, ``Explaining explanations in ai,''
  in {\em Proceedings of the conference on fairness, accountability, and
  transparency}, pp.~279--288, 2019.

\bibitem{kment2006counterfactuals}
B.~Kment, ``Counterfactuals and explanation,'' {\em Mind}, vol.~115, no.~458,
  pp.~261--310, 2006.

\bibitem{wang2019designing}
D.~Wang, Q.~Yang, A.~Abdul, and B.~Y. Lim, ``Designing theory-driven
  user-centric explainable ai,'' in {\em Proceedings of the CHI Conference on
  Human Factors in Computing Systems}, pp.~1--15, 2019.

\bibitem{zhu2018explainable}
J.~Zhu, A.~Liapis, S.~Risi, R.~Bidarra, and G.~M. Youngblood, ``Explainable
  {AI} for designers: A human-centered perspective on mixed-initiative
  co-creation,'' in {\em IEEE Conference on Computational Intelligence and
  Games (CIG)}, pp.~1--8, 2018.

\bibitem{hoffman2017explaining}
R.~R. Hoffman and G.~Klein, ``{Explaining explanation, part 1: Theoretical
  foundations},'' {\em IEEE Intelligent Systems}, vol.~32, no.~3, pp.~68--73,
  2017.

\bibitem{van1994think}
M.~Van~Someren, Y.~Barnard, and J.~Sandberg, ``The think aloud method: a
  practical approach to modelling cognitive,'' {\em London: AcademicPress},
  1994.

\bibitem{kim2018textual}
J.~Kim, A.~Rohrbach, T.~Darrell, J.~Canny, and Z.~Akata, ``Textual explanations
  for self-driving vehicles,'' in {\em Proceedings of the European conference
  on computer vision (ECCV)}, pp.~563--578, 2018.

\bibitem{xu2020explainable}
Y.~Xu, X.~Yang, L.~Gong, H.-C. Lin, T.-Y. Wu, Y.~Li, and N.~Vasconcelos,
  ``Explainable object-induced action decision for autonomous vehicles,'' in
  {\em Proceedings of the IEEE/CVF Conference on Computer Vision and Pattern
  Recognition}, pp.~9523--9532, 2020.

\bibitem{omeiza2021towards}
D.~Omeiza, H.~Webb, M.~Jirotka, and L.~Kunze, ``Towards accountability:
  providing intelligible explanations in autonomous driving,'' in {\em
  Proceedings of the IEEE Symposium on Intelligent Vehicle}, IEEE, 2021.

\bibitem{nahata2021assessing}
R.~Nahata, D.~Omeiza, R.~Howard, and L.~Kunze, ``Assessing and explaining
  collision risk in dynamic environments for autonomous driving safety,'' 2021.

\bibitem{schneider2021explain}
T.~Schneider, J.~Hois, A.~Rosenstein, S.~Ghellal, D.~Theofanou-F{\"u}lbier, and
  A.~R. Gerlicher, ``Explain yourself! transparency for positive ux in
  autonomous driving,'' in {\em Proceedings of the 2021 CHI Conference on Human
  Factors in Computing Systems}, pp.~1--12, 2021.

\bibitem{koo2015did}
J.~Koo, J.~Kwac, W.~Ju, M.~Steinert, L.~Leifer, and C.~Nass, ``Why did my car
  just do that? explaining semi-autonomous driving actions to improve driver
  understanding, trust, and performance,'' {\em International Journal on
  Interactive Design and Manufacturing (IJIDeM)}, vol.~9, no.~4, pp.~269--275,
  2015.

\bibitem{ramanishka2018toward}
V.~Ramanishka, Y.-T. Chen, T.~Misu, and K.~Saenko, ``{Toward driving scene
  understanding: A dataset for learning driver behavior and causal
  reasoning},'' in {\em Proceedings of the IEEE Conference on Computer Vision
  and Pattern Recognition}, pp.~7699--7707, 2018.

\bibitem{stepin2021factual}
I.~Stepin, A.~Catala, M.~Pereira-Fari{\~n}a, and J.~M. Alonso, ``Factual and
  counterfactual explanation of fuzzy information granules,'' {\em
  Interpretable Artificial Intelligence: A Perspective of Granular Computing},
  vol.~937, p.~153, 2021.

\bibitem{anjomshoae2021context}
S.~Anjomshoae, D.~Omeiza, and L.~Jiang, ``Context-based image explanations for
  deep neural networks,'' {\em Image and Vision Computing}, p.~104310, 2021.

\bibitem{palczewska2013interpreting}
A.~Palczewska, J.~Palczewski, R.~M. Robinson, and D.~Neagu, ``Interpreting
  random forest models using a feature contribution method,'' in {\em 2013 IEEE
  14th International Conference on Information Reuse \& Integration (IRI)},
  pp.~112--119, IEEE, 2013.

\bibitem{singh2021road}
G.~Singh, S.~Akrigg, M.~Di~Maio, V.~Fontana, R.~J. Alitappeh, S.~Saha,
  K.~Jeddisaravi, F.~Yousefi, J.~Culley, T.~Nicholson, {\em et~al.}, ``Road:
  The road event awareness dataset for autonomous driving,'' {\em arXiv
  preprint arXiv:2102.11585}, 2021.

\bibitem{de2017people}
M.~M. De~Graaf and B.~F. Malle, ``How people explain action (and autonomous
  intelligent systems should too),'' in {\em 2017 AAAI Fall Symposium Series},
  2017.

\bibitem{nisbett1977telling}
R.~E. Nisbett and T.~D. Wilson, ``Telling more than we can know: Verbal reports
  on mental processes.,'' {\em Psychological review}, vol.~84, no.~3, p.~231,
  1977.

\bibitem{kahneman2009conditions}
D.~Kahneman and G.~Klein, ``Conditions for intuitive expertise: a failure to
  disagree.,'' {\em American psychologist}, vol.~64, no.~6, p.~515, 2009.

\end{thebibliography}
}


\end{document}